\documentclass[10pt,twocolumn,letterpaper]{article}

\usepackage{iccv}
\usepackage{times}
\usepackage{epsfig}
\usepackage{graphicx}
\usepackage{amsmath}
\usepackage{amssymb}
\usepackage{mathrsfs}
\usepackage{array}

\usepackage[pagebackref=false,breaklinks=true,letterpaper=true,colorlinks,bookmarks=false]{hyperref}

\iccvfinalcopy 

\def\iccvPaperID{1917} 

\ificcvfinal\fi
\begin{document}

\title{`Skimming-Perusal' Tracking: A Framework for Real-Time and Robust Long-term Tracking}

\author{Bin Yan$^{1\dag}$, Haojie Zhao$^{1\dag}$, Dong Wang$^{1}\thanks{Corresponding Author: Dr. Dong Wang, wdice@dlut.edu.cn}$,  Huchuan Lu$^1$ and Xiaoyun Yang$^2$\\
$^1$School of Information and Communication Engineering, Dalian University of Technology, China\\
$^2$China Science IntelliCloud Technology Co., Ltd., \quad
$^{\dag}$ Equal Contribution\\
{\tt\footnotesize \{yan\_bin, haojie\_zhao\}@mail.dlut.edu.cn, 
\{wdice, lhchuan\}@dlut.edu.cn, xiaoyun.yang@intellicloud.ai}
}

\maketitle
\ificcvfinal\thispagestyle{empty}\fi

\begin{abstract}
Compared with traditional short-term tracking, long-term tracking poses more 
challenges and is much closer to realistic applications. However, few works 
have been done and their performance have also been limited. 
In this work, we present a novel robust and real-time long-term tracking framework 
based on the proposed skimming and perusal modules. The perusal module consists 
of an effective bounding box regressor to generate a series of candidate proposals 
and a robust target verifier to infer the optimal candidate with its confidence score. 
Based on this score, our tracker determines whether the tracked object being present 
or absent, and then chooses the tracking strategies of local search or global search 
respectively in the next frame.
To speed up the image-wide global search, a novel skimming module is designed to 
efficiently choose the most possible regions from a large number of sliding windows. 
Numerous experimental results on the VOT-2018 long-term and OxUvA long-term 
benchmarks demonstrate that the proposed method achieves the best performance 
and runs in real-time. 
The source codes are available at \href{https://github.com/iiau-tracker/SPLT}
{https://github.com/iiau-tracker/SPLT}.
\end{abstract}

\section{Introduction}
Online visual tracking is one of most important problems in computer vision, and has 
many practical applications including video surveillance, behavior analysis, visual 
navigation, augmented reality and so on. It is a very tough task to design a robust and 
efficient tracker caused by the challenges from both foreground and background variations. 
These challenges include occlusion, illumination variation, viewpoint change, rotation, 
motion blur, to name a few. 
Due to the breakthrough of deep learning and the construction of large-scale benchmarks, 
numerous trackers~\cite{ECO,MDNet,LSART,DRT,SiameseRPN,DSiam,RTMDNet,DVT-Review,ATOM,SiamMask,SiamRPNplus,ASRCF,CascadedSiameseRPN,GradNet,CPF-TIP19,MCPF-TPAMI19} 
have recently achieved very promising performance. 

However, most of existing trackers and datasets focus on the short-term tracking task, 
the setting of which is that the tracked object is almost always in the camera filed of view 
but not necessarily fully visible. 
There still exists a big gap between the short-term tracking setting and the realistic tracking 
applications. 
In recent, some researchers have focused on the long-term tracking task and attempted to 
create related large-scale benchmarks (such as VOT2018LT~\cite{VOTLT} and 
OxUvA~\cite{OxUvA}). 

Compared with short-term tracking, the long-term tracking task additionally requires the 
tracker having the capability to capture the tracked object in long-term videos and to handle 
the frequent target disappearance and reappearance. Thus, it poses more challenges than 
short-term tracking  mainly from two aspects. 
First, the frame length in long-term datasets is much greater than that in short-term scenarios. 
For examples, the average numbers of frame length in the VOT2018LT~\cite{VOTLT}, 
OxUvA~\cite{OxUvA}, OTB2015~\cite{OTB2015} and VOT2018~\cite{VOT2018report} 
benchmarks are $4196$, $4235$, $590$ and $350$, respectively (the former two are long-term 
datasets and the latter two are well-known short-term datasets). Second, there exist a large 
number of \emph{absent} labels in long-term tracking datasets.  
Thus, it is critical for long-term trackers to capture the tracked object in long-term sequences, 
determine whether the target is \emph{present} or \emph{absent}, and have the capability of 
image-wide re-detection. 

Until now, some long-term trackers have been developed based on hand-crafted features, 
including TLD~\cite{TLD}, LCT~\cite{LCT}, FCLT~\cite{FCLT}, MUSTer~\cite{MUSTer}, 
CMT~\cite{CMT} and EBT~\cite{EBT}. Although these methods obtain some achievements 
on a small number of long-term videos, they cannot achieve satisfactory performance on recent 
long-term benchmarks (see experiments in~\cite{VOT2018report,OxUvA}). 
Recently, some deep-learning-based algorithms~\cite{PTAV,VOT2018report,MBMD} are proposed 
for long-term tracking and significantly improved the tracking performance. 
But there still lacks of a robust and real-time framework to address the long-term tracking task. 

In this work, we propose a novel `Skimming-Perusal' tracking framework for long-term tracking. 
The perusal module aims to precisely capture the tracked object in a local search region; while the 
skimming module focuses on efficiently selecting the most possible candidate regions and 
significantly speeding up the image-wide re-detection process.
Our main contributions can be summarized as follows. 
\vspace{-6mm}
\begin{itemize}
\setlength{\itemsep}{0pt}
\setlength{\parsep}{0pt}
\setlength{\parskip}{0pt}
\item \emph{A novel `Skimming-Perusal' framework based on deep networks is proposed 
to address the long-term tracking task. Both skimming and persual modules are offline trained 
and directly used during the tracking process. Our framework is simple yet effective, which 
could be served as a new baseline for long-term tracking.}
\item \emph{A novel perusal module is developed to precisely capture the tracked object in 
a local search region, which is comprised of an effective bounding box regressor based on 
SiameseRPN and a robust offline-trained verifier based on deep feature embedding.} 
\item \emph{A novel skimming module is designed to efficiently select the most possible 
local regions from densely sampled sliding windows, which could speed up the image-wide 
re-detection process when the target is absent.}
\item \emph{Numerous experimental results on the VOT2018LT and OxUvA datasets 
show that our tracker achieves the best accuracies with real-time performance.}
\end{itemize}

\vspace{-4mm}
\section{Related Work}
\vspace{-1mm}
{\flushleft \textbf{Traditional Long-term Tracking.}} In~\cite{TLD}, Kalal~\emph{et al.} 
propose a tracking-learning-detection (TLD) algorithm for long-term tracking, which exploits 
an optical-flow-based matcher for local search and an ensemble of weak classifiers for global 
re-detection. Following the idea of TLD, Ma~\emph{et al.}~\cite{LCT} develop a long-term 
correlation tracker (LCT) using a KCF method as a local tracker and a random ferns classifier 
as a detector.  The fully correlational long-term tracker (FCLT)~\cite{FCLT} maintains several 
correlation filters trained on different time scales as a detector and exploits the correlation 
response to guide the dynamic interaction between the short-term tracker and long-term detector. 

Besides, some researchers have addressed the long-term tracking task using the keypoint 
matching or global proposal scheme. 
The CMT~\cite{CMT} method utilizes a keypoint-based model to conduct long-term tracking, 
and the MUSTer~\cite{MUSTer} tracker exploits an integrated correlation filter for short-term 
localization and a keypoint-based matcher for long-term tracking. But the keypoint extractors 
and descriptors are often not stable in complicated scenes. 
In~\cite{EBT},  Zhu~\emph{et al.} develop an EdgeBox Tracking (EBT) method to generate 
a series of candidate proposals using EdgeBox~\cite{Zitnick-ECCV2014-EdgeBox} and verify 
these proposals using structured SVM~\cite{Hare-SSVM} with multi-scale color histograms. 
However, the edge-based object proposal is inevitably susceptible to illumination variation 
and motion blur.
The above-mentioned trackers have attempted to address long-term tracking from different 
perspectives, but their performance are not satisfactory since they merely exploit hand-crafted 
low-level features. 
In this work, we develop a simple yet effective long-term tracking framework based on deep 
learning, whose goal is to achieve high accuracy with real-time performance.

\vspace{-5mm}
{\flushleft \textbf{Deep Long-term Tracking.}} Recently, some researchers have attempted 
to exploit deep-learning-based models for long-term tracking. Fan~\emph{et al.}~\cite{PTAV} 
propose a parallel tracking and verifying (PTAV) framework, which effectively integrates a 
real-time tracker and a high accurate verifier for robust tracking. The PTAV method performs 
much better than other compared trackers on the UAV20L dataset. 
Valmadre~\emph{et al.}~\cite{OxUvA} implement a long-term tracker, named as SiamFC+R. 
This method equips SiamFC~\cite{SiameseFC} with a simple re-detection scheme, and finds 
the tracked object within a random search region when the maximum score of the SiamFC's 
response is lower than a given threshold. The experimental results demonstrate that the 
SiamFC+R tracker achieves significantly better performance than the original SiamFC method 
on the OxUvA dataset. But the SiamFC's score map is not always reliable, which limits the 
performance of the SiamFC+R tracker. 

In~\cite{MBMD}, Zhang~\emph{et al.} combine an offline-trained bounding box regression network 
and an online-learned verification network to design a long-term tracking framework. 
The regression network determines the bounding boxes of the tracked object in a search region; 
while the verification network verifies whether the tracking result is reliable or not and makes 
the tracker dynamically switch between local search and global search states. The global 
search explicitly conducts image-wide re-detection by exploiting a sliding window strategy in 
the entire image. 
This method provides a complete long-term tracking framework, and also is the winner of 
the VOT2018 long-term challenge. 
However, the adopted straightforward sliding window strategy and online-learned verification 
model make it be very slow and far from the real-time applications (merely 2.7fps reported 
in~\cite{MBMD} and 4.4fps in our experiment setting).
In~\cite{VOT2018report}, Zhu~\emph{et al.} propose a DaSiam\_LT method for long-term tracking, 
which extends the original SiameseRPN tracker~\cite{SiameseRPN} by introducing a local-to-global 
search region strategy. During the tracking process, the size of the search region will be iteratively 
increasing when the tracking failure is indicated. The distractor-aware training and 
inference~\cite{DSiam} make the 
output tracking scores suitable to determine whether the tracker fails or not. 
The DaSiam\_LT tracker achieves the second best performance in the VOT2018 long-term 
challenge, however, it requires very large numbers of image sequences for offline training. 
Besides, the DaSiam\_LT method does not explicitly include the image-wide re-detection 
scheme, thereby failing in the re-detection experiment. 
Compared with aforementioned methods, our goal is to develop a simple but effective long-term 
tracking framework with high accuracy and real-time performance. 

\section{`Skimming-Perusal' Tracking Framework}
In this work, we propose a novel `Skimming-Perusal' framework to address the long-term tracking 
problem. There exist two fundamental modules: skimming and perusal. The perusal module aims 
to conduct robust object regression and verification in a local search region; while the skimming 
module focuses on quickly selecting the most possible candidate regions within a large number of 
sliding windows when the tracker runs in the global search state. 

\begin{figure}[!h]
\begin{center}
\includegraphics[width=1.0\linewidth,height=0.4\linewidth]{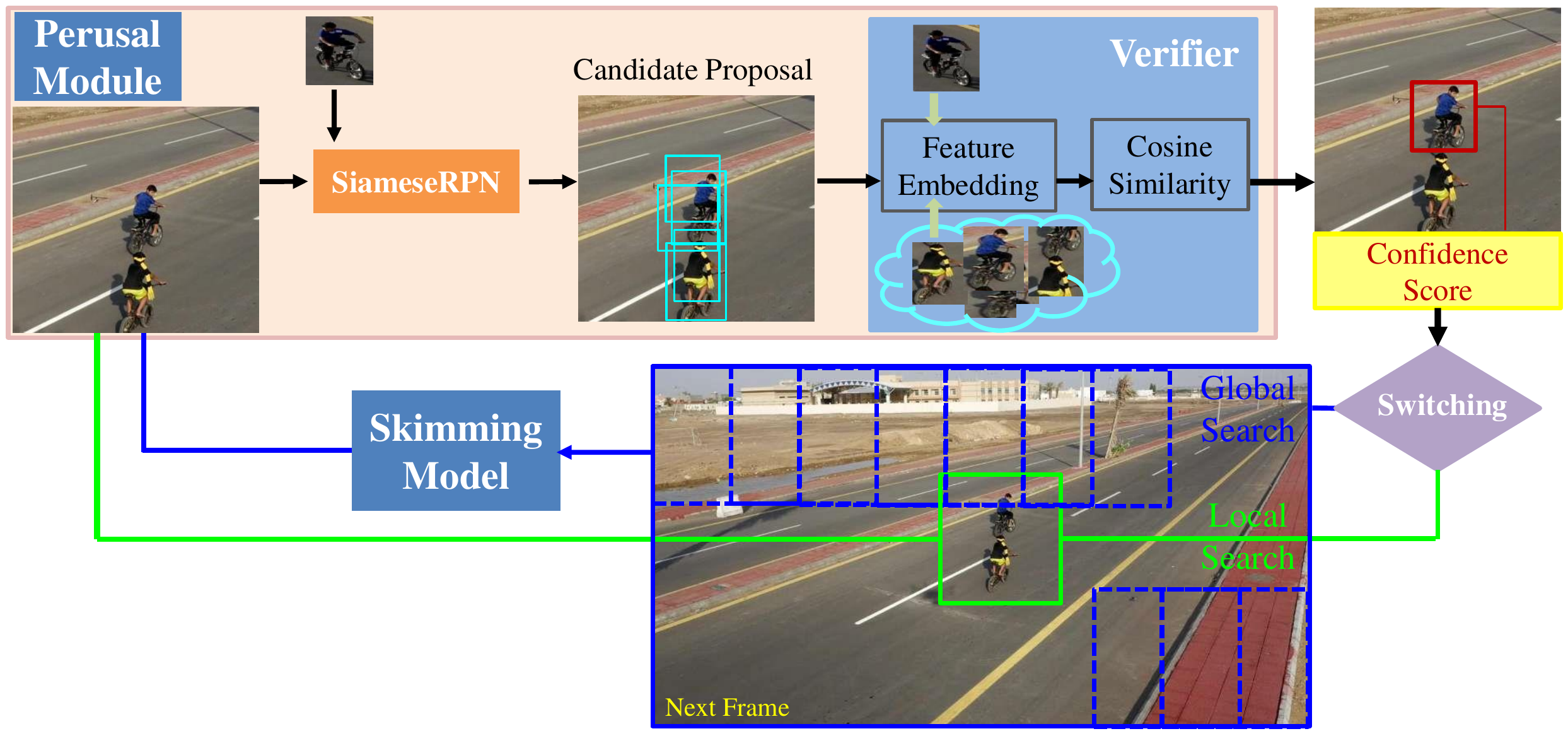}
\end{center}
\vspace{-3mm}
\caption{Our `Skimming-Perusal' long-term tracking framework. Better viewed in color with zoom-in.}
\label{fig-framework}
\end{figure}

The overall framework is presented in Figure~\ref{fig-framework}. Our tracker first searches the 
target in a local search region (four times of the target size) using the perusal module. After obtaining 
the best candidate in each frame, our tracker treats the tracked object as \emph{present} or \emph{absent} 
based on its confidence score and then determines the search state (local  search or global search) 
in the next frame. 
If the confidence score is higher than a pre-defined threshold, the tracker treats the target as \emph{present} 
and continues to track the target in the local search region centered by the object location. 
Otherwise, the tracker regards the target as \emph{absent} and conducts global search in the next frame. 
To be specific, our global search scheme crops a series of local search regions using sliding windows 
and then deals with these regions using the perusal module.
To speed it up, we develop a novel skimming module to efficiently select the most likely local regions 
and effectively handle these regions using the perusal module. 
The detailed descriptions of our skimming and perusal modules are presented as follows.

\subsection{Robust Local Perusal with Offline-learned Regression and Verification Networks}
\begin{figure}[!h]
\begin{center}
\includegraphics[width=1.0\linewidth,height=0.33\linewidth]{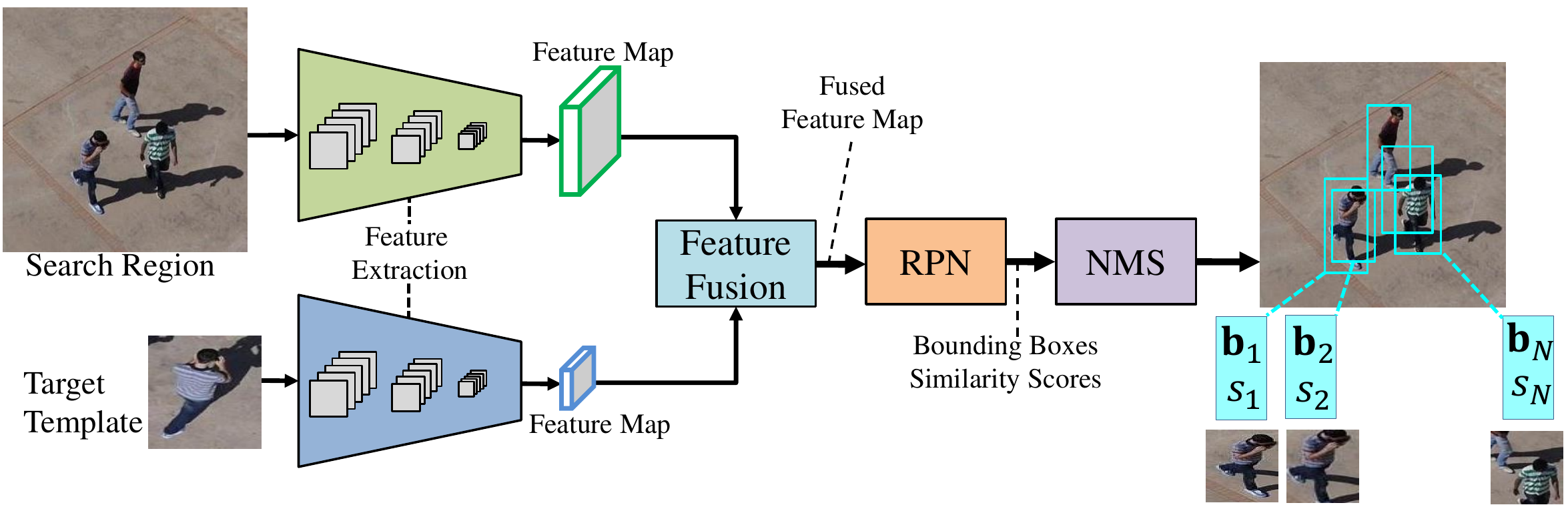}
\end{center}
\vspace{-3mm}
\caption{The adopted SiameseRPN module in our framework. Better viewed in color with zoom-in.}
\label{fig-SiameseRPN}
\end{figure}

Our perusal module is composed of an offline-learned SiameseRPN model and an offline-learned 
verification model (shown in Figure~\ref{fig-framework}). 
The former one generates a series of candidate proposals within a local search region, and the latter 
one verifies them and determines the best candidate. 
\vspace{-2mm}
{\flushleft \textbf{SiameseRPN.}} The SiameseRPN method~\cite{SiameseRPN} improves the 
the classical SiamFC method~\cite{SiameseFC} by introducing a region proposal network, which 
allows the tracker to estimate the bounding box of variable aspect ratio effectively. 
In this work, we choose SiameseRPN as our basic regressor due to its robustness and efficiency. 
To be specific, we adopt a variant of the original SiameseRPN method proposed in~\cite{MBMD} 
since its training and testing codes are all publicly available. The flowchart of the adopted SiameseRPN 
model is illustrated in Figure~\ref{fig-SiameseRPN}.
Given a target template $\mathcal{Z}$ and a local search region $\mathcal{X}$, the SiameseRPN model 
generates a set of bounding boxes ${\bf{B}} = \left[ {{{\bf{b}}_1},{{\bf{b}}_2},...,{{\bf{b}}_N}} 
\right]$ with their corresponding similarity scores ${\bf{s}} = \left[ {{s_1},{s_2},...,{s_N}} \right]$ ($N$ 
denotes the total number of candidates). 

\begin{figure*}[!t]
\begin{center}
\begin{tabular}{ccc}
\includegraphics[width=0.32\linewidth,height=0.26\linewidth]{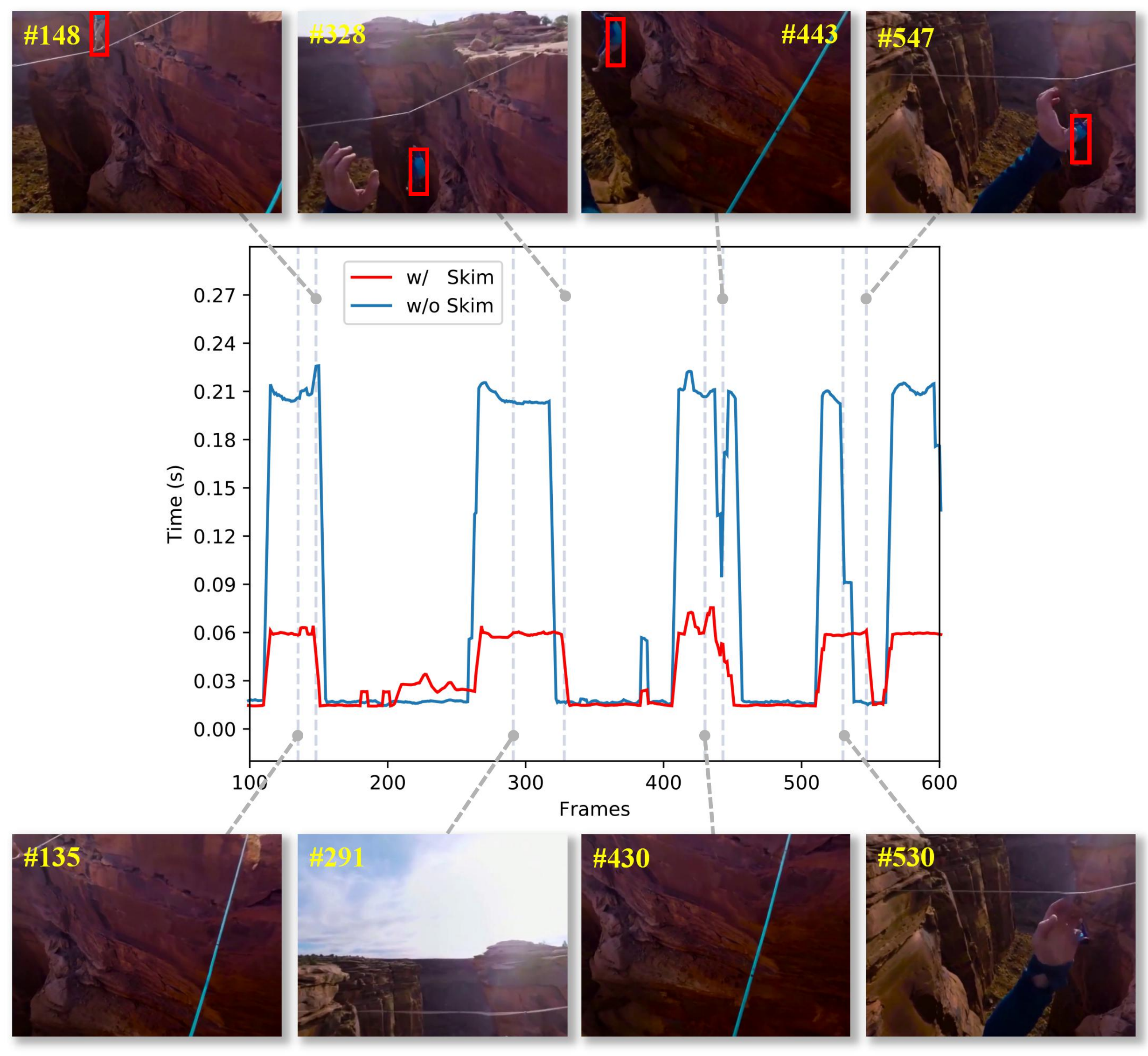} \ &
\includegraphics[width=0.32\linewidth,height=0.26\linewidth]{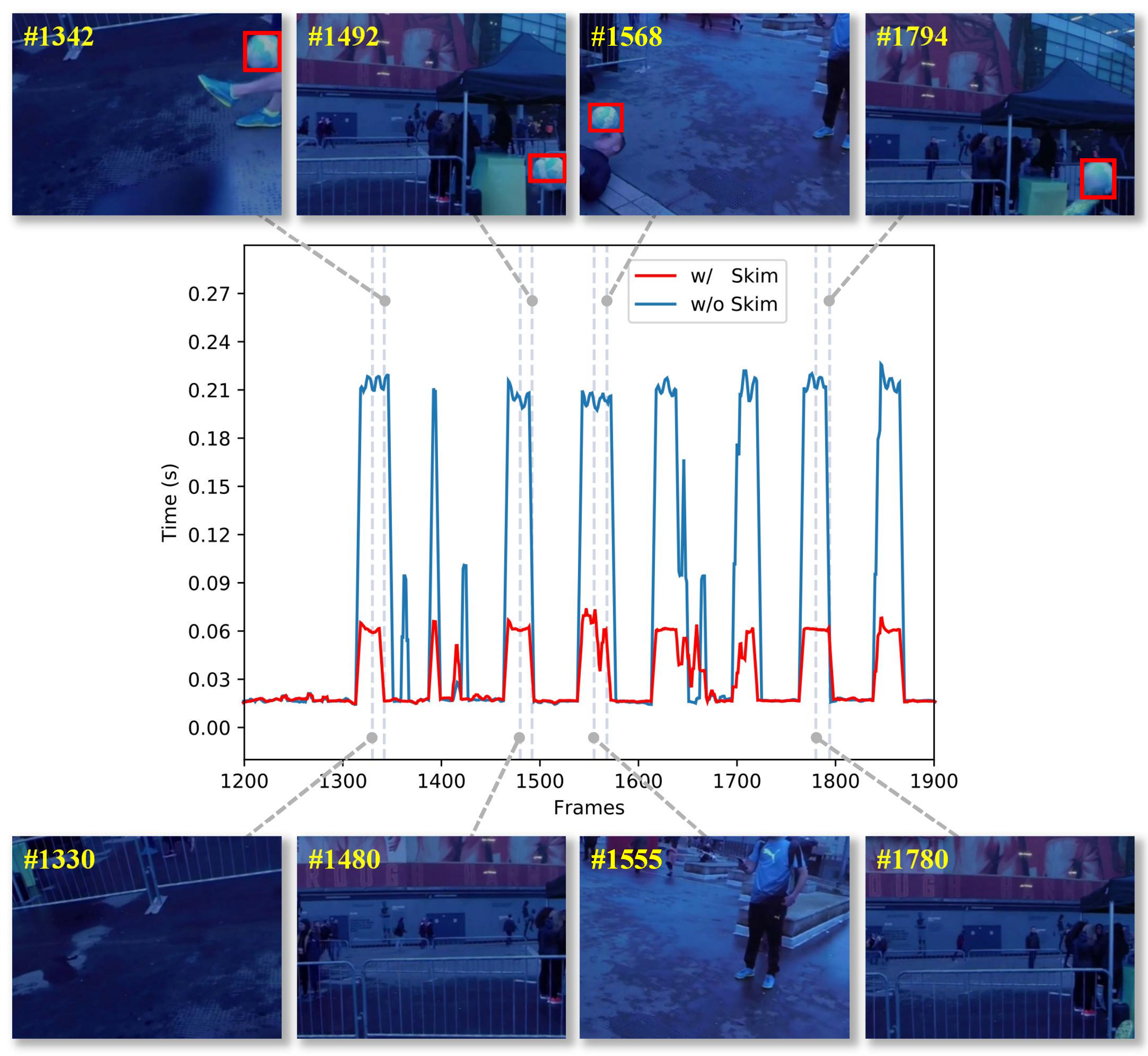} \ &
\includegraphics[width=0.32\linewidth,height=0.26\linewidth]{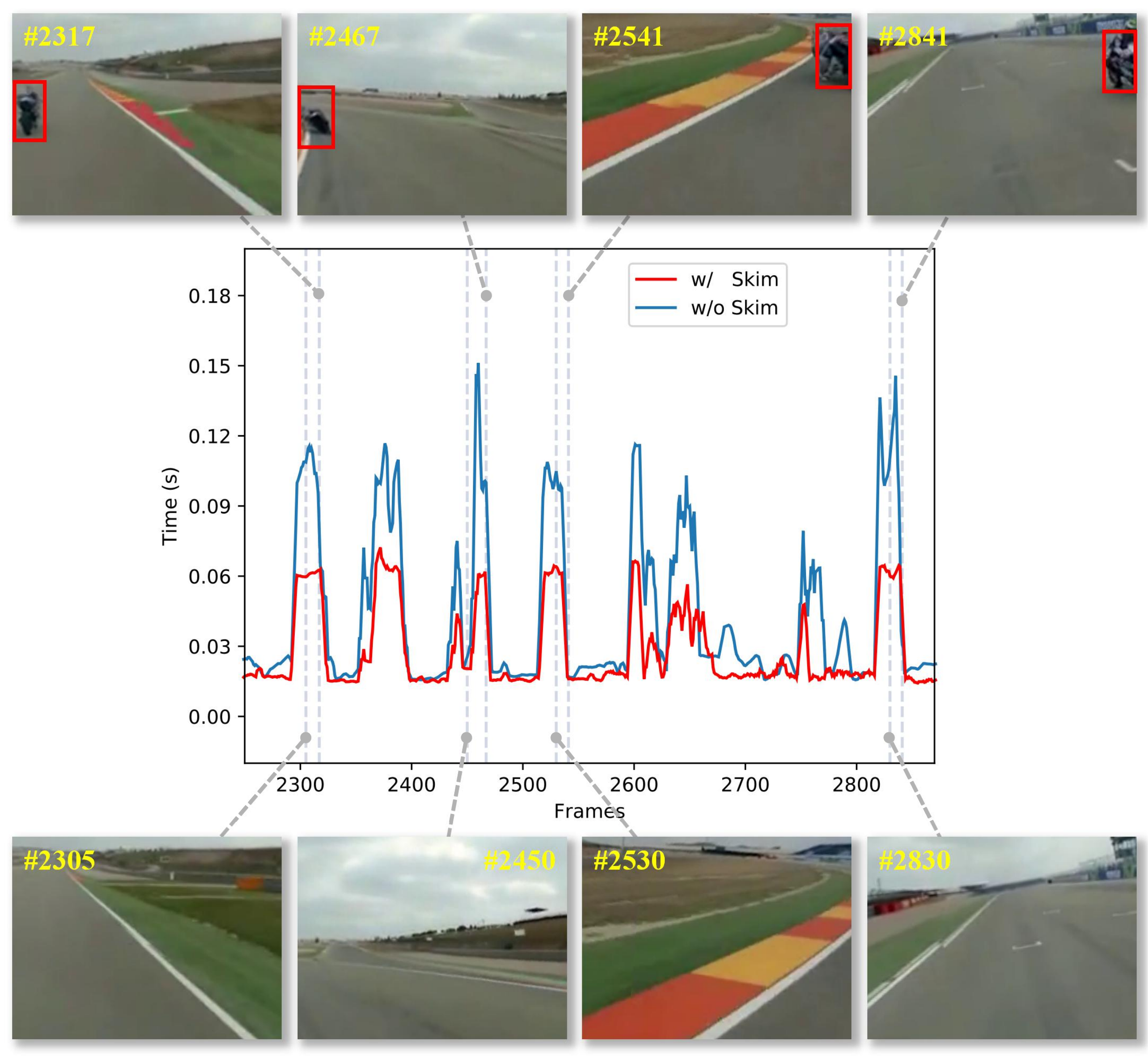}\\
(a) \footnotesize{tightrope} & (b)  \footnotesize{freestyle} & (c)  \footnotesize{yamaha}\\
\end{tabular}
\end{center}
\vspace{-3mm}
\caption{Illustration of the effectiveness of the proposed skimming module. The abbreviations 
`w/ skim' and `w/o skim' denote our trackers with and without the skimming module, respectively.}
\label{fig-skim-exp}
\end{figure*}

Simply, we can directly locate the tracked object based on the bounding box with the highest score 
in the current frame (i.e., obtain the optimal bounding box as ${{\bf{b}}_{{i^*}}},{i^*} = 
\arg {\max _i}\left\{ {{s_i}} \right\}$. 
However, this manner makes the tracker unstable and easily drift to some distractors.
This is mainly attributed to the multi-task learning manner in SiameseRPN. The joint learning 
bounding box predictions and classification scores usually generate accurate box proposals 
but unreliable scores for the tracking task.  Thus, we merely exploit SiameseRPN to generate 
candidate proposals, and then infer their confidence score using an additional verifier.

\vspace{-2mm}
{\flushleft \textbf{Offline-learned Verification Network.}} To ensure the efficiency of our tracking 
framework, we attempt to exploit an offline-trained verifier based on deep feature embedding~\cite{FaceNet}. 
Specifically, we learn an embedding function $f\left( . \right)$ to embed the target template and the 
candidate proposals into a discriminative Euclidean space.  The discriminative ability is ensured 
by the following triplet loss, 
\begin{equation}
{\sum\limits_{i = 1}^M {\left[ {\left\| {f\left( {\mathcal{Y}_i^a} \right) - f\left( {\mathcal{Y}_i^p} \right)} 
\right\|_2^2 - \left\| {f\left( {\mathcal{Y}_i^a} \right) - f\left( {\mathcal{Y}_i^n} \right)} \right\|_2^2 + \alpha } 
\right]} _ + }, 
\end{equation}
where ${\mathcal{Y}_i^a}$ denotes the $i$-th anchor of a specific target, ${\mathcal{Y}_i^p}$ is a positive 
sample (i.e., one of other images of the target), and ${\mathcal{Y}_i^n}$ is a negative sample of any other target 
or background. $\alpha$ is a margin value (simply set to $0.2$ in this work). $\mathcal{T}$ is the set of all possible 
triplet pairs in the training set and has cardinality $M$. The construction of training triplets and the hyper-parameters 
are presented in Section 3.3.

During tracking, we can determine the confidence scores $\left[ {{c_1},{c_2},...,{c_N}} \right]$ of the 
candidate proposals $\left[ {{{\bf{b}}_1},{{\bf{b}}_2},...,{{\bf{b}}_N}} \right]$ using a thresholding 
cosine similarity metric as 
\begin{equation}
{c_i} = \max \left( {\frac{{{f^ \top }\left( \mathcal{Z} \right)f\left( {\phi \left( {{{\bf{b}}_i}} \right)} \right)}}
{{{{\left\| {f\left( \mathcal{Z} \right)} \right\|}_2}{{\left\| {f\left( {\phi \left( {{{\bf{b}}_i}} \right)} \right)} 
\right\|}_2}}},0} \right), 
\label{eq-confidence}
\end{equation}
where $\mathcal{Z}$ is the target template in the first frame and ${\phi \left( {{{\bf{b}}_i}} \right)}$ denotes 
the image cropped within the $i$-th bounding box ${{{\bf{b}}_i}}$. 

After that, the optimal candidate can be determined as ${{\bf{b}}_{{i^*}}}$ (${i^*} = \arg {\max _i}
\left\{ {{c_i}} \right\}$), whose corresponding confidence score is ${c_{{i^*}}}$.  
Finally, we exploit a threshold-based switching strategy to make interactions between 
local search and global search dynamically.  If the score ${c_{{i^*}}}$ is larger than 
a pre-defined threshold $\theta$, our tracker treats the target being \emph{present} and 
continues to conduct local search in the next frame. Otherwise, it considers the tracked 
object as \emph{absent} and then invokes the global search scheme in the next frame. 
$\theta$ is set to $0.65$ in this work and discussed in Section 4.2.

\vspace{-2mm}
{\flushleft \textbf{Cascaded Training.}} To enhance the performance of our perusal module, 
we exploit a cascaded training strategy to make our verifier more robust. First, we 
train the SiameseRPN and verification network individually. 
Then, we apply the perusal module to the training set and collect the mis-classified samples as 
hard examples. 
Finally, we fine-tune the verification network using the collected hard examples. The training 
details are presented in Section 3.3.

\subsection{Efficient Global Search with Offline-learned Skimming Module}
The long-term tracker usually combines a local tracker and a global re-detector, 
and invokes the global re-detector when the object is \emph{absent}. The sidling 
window technique is widely used to conduct global search~\cite{LCT,MBMD}, 
by which the local region in each window is utilized to determine whether the 
object is \emph{present} or not. However, this manner is very time-consuming 
especially when deep-learning-based models are used. For example, the recent 
MBMD tracker~\cite{MBMD} (the winner of VOT2018 long-term challenge) 
merely runs less than 5fps (very far from real-time performance). 

To address this issue, we propose a skimming module to conduct fast global search by 
efficiently selecting the most possible candidate regions from a large number of sliding 
windows. 
Figure~\ref{fig-skim-exp} demonstrates some representative examples, from which 
we can see that our skimming scheme could significantly reduce the running time 
when the tracker conducts image-wide re-detection (i.e., the time interval between target 
disappearance and reappearance). 

\begin{figure}[!h]
\begin{center}
\includegraphics[width=1.0\linewidth,height=0.4\linewidth]{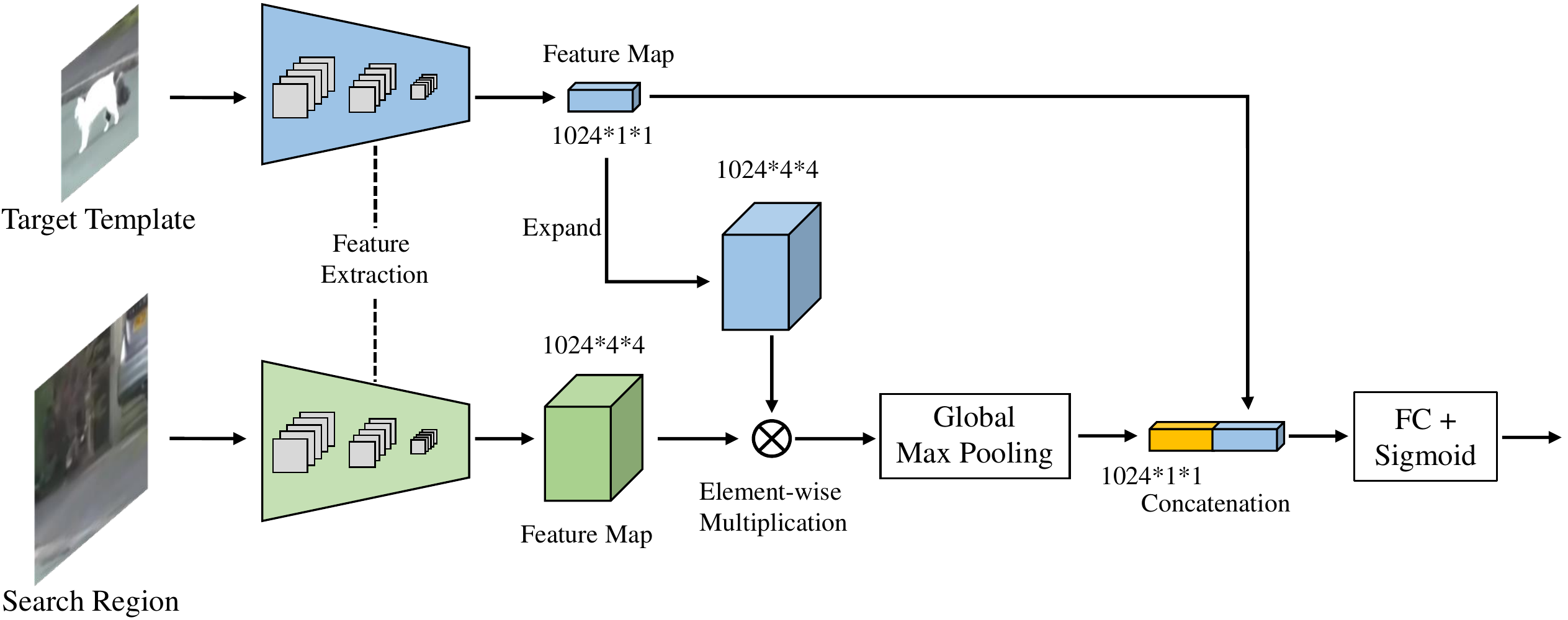}
\end{center}
\vspace{-3mm}
\caption{The network architecture of our skimming module. Better viewed in color with zoom-in.}
\label{fig-skim}
\end{figure}

Given a target template $\mathcal{Z}$ and a search region $\mathcal{X}$, the skimming 
module aims to learn a function $p = g\left( {\mathcal{Z},\mathcal{X}} \right)$, where 
$p$ indicates whether the target appears in this region or not.
The function $g\left( {.,.} \right)$ is implemented using deep convolutional neural networks 
(CNN), whose network architecture is presented in Figure~\ref{fig-skim}. 
Both target template and search region are fed into CNN feature extractors, and then their 
feature maps are fused and concatenated into a long vector. Finally, the fully connected (FC) 
layer with the sigmoid function is added to conduct a binary classification. 
The cross entropy loss is adopted to train this network. 

When the tracker runs on global search, a series of sliding windows are densely sampled. 
We first apply our skimming module on these regions, and then select 
the top-$K$ candidates based on their classification margins and discard the remaining ones as 
distractors. Only selected regions will be further handled using the perusal module (SiameseRPN+Verifier), 
which makes our tracker very efficient in image-wide re-detection. In addition, our skimming module 
could improve the tracker's robustness since it filters out some distractors and alleviates tracking drift. 
The parameter $K$ is set as $3$ in this work.

\subsection{Implementation Details}
In this subsection, we present some key implementation details due to space limitation. 
The detailed parameter settings can be found in our source codes. 
\vspace{-2mm}
{\flushleft \textbf{Network Architectures and Training Dataset}:}
In this work, our regression and skimming models have similar structures. First, we use 
MobileNet(V1)~\cite{MobileNets} as our feature extractor. Second, we downsample the 
spatial resolution of the template feature to $1 \times 1$ using average pooling. 
Besides, the Siamese branches do not share parameters. For the verification model, 
we adopt ResNet50 as the backbone of our verifier. 
The resolution of the template image and candidate proposals is resized to $127 \times 127$, 
and the resolution of the search regions is resized to $300 \times 300$. 
The parameters of aforementioned networks are initialized with the ImageNet pre-trained 
models and then fine-tuned on the ImageNet DET and VID datasets~\cite{ImageNet}. 
\vspace{-2mm}
{\flushleft \textbf{Training Data Preparation}:}
In order to make our regression module obtain capability to regress any kind of object for a given object template 
as well as learn a generic matching function for tracking to tolerate the common appearance variations, we combine 
the ImageNet DET and VID datasets, and introduce some data augmentations like horizontally flipping and 
random erasing~\cite{RandomErasing}. 
For the verification network, theoretically we should choose hard examples as triplet pairs to speed up convergence 
and boost the discriminative power. To achieve this goal, we adopt the following sampling strategy: (1) randomly 
choosing one video from training set and picking its initial target patch as the anchor; (2) stochastically choosing one 
frame within this video as the positive one; and (3) randomly choosing another frame from the video belonging to a 
different object class as the negative one. 
Besides, we exploit the cascaded training strategy (see Section 3.1) to further mine hard samples and fine-tune the 
verification network.
For the skimming network, we randomly crop search regions around the target as positive samples. In order to obtain 
negative samples more conveniently, we directly mask the target with mean pixel value, then crop negative samples 
using the same strategy as cropping positive ones. The template generation manner is consistent with the SiamFC 
method~\cite{SiameseFC}.
\vspace{-2mm}
{\flushleft \textbf{Training Strategy: }}
The regression, verification and skimming modules are trained independently in an end-to-end 
fashion using the batch gradient descent optimizer with the  momentum of 0.9.
The batch sizes of regression and skimming networks are set to $32$, and the batch size of the verification 
network is chosen as $64$. 
We train regression and skimming networks for 500,000 iterations and $20$ epochs 
with the same learning rate of ${\rm{1}}{e^{ - 3}}$, respectively.  
For the verification network, we train it for $70$ epochs and exploit 60,000 triplet pairs in each epoch. 
The learning rate is initially set to ${\rm{1}}{e^{ - 2}}$ and gradually decayed to its ${1 \mathord{\left/
   {\vphantom {1 {10}}} \right.\kern-\nulldelimiterspace} {10}}$ every $20$ epochs.

\section{Experiments}
In this work, we implement our tracker using Python with the Tensorflow~\cite{Tensorflow} 
and Keras deep learning libraries. The proposed method is tested on a PC machine with an 
Inter i7 CPU (32G RAM) and a NVIDIA GTX1080Ti GPU (11G memory), which runs in 
real-time with $\textbf{25.7}$ frames per second (fps). Our tracker is denoted as 
$\textbf{SPLT}$. 
Both training and testing codes are available at \href{https://github.com/iiau-tracker/SPLT}
{https://github.com/iiau-tracker/SPLT}.

We compare our tracker with other competing algorithms on the VOT-2018 long-term 
(VOT2018LT) dataset~\cite{VOTLT} and OxUvA long-term dataset~\cite{OxUvA}. 
The quantitative evaluations and ablation studies are reported as follows. 

\subsection{Results on \textbf{VOT2018LT}}

The VOT2018LT~\cite{VOTLT} dataset is first presented in Visual Object 
Tracking (VOT) challenge 2018 to evaluate the performance of different long-term 
trackers. This dataset includes $35$ sequences of various objects (e.g., persons, 
cars, bicycles and animals), with the total frame length of $146847$ frames and 
the resolution ranges between $1280\times720$ and $290\times217$. The 
construction of this dataset fully takes object disappearance into account, where 
each sequence contains on average $12$ long-term object disappearances and each 
disappearance lasts on average $40$ frames. The ground truths of the tracked objects 
are annotated by bounding boxes, and sequences are annotated by nine visual attributes 
(including full occlusion, out-of-view, partial occlusion, camera motion, fast motion, 
scale change, aspect ratio change, viewpoint change and similar objects). 

The evaluation protocol of the VOT2018LT~\cite{VOTLT} dataset includes two 
aspects: accuracy evaluation and re-detection evaluation. 
First, the accuracy evaluation measures the performance of a given long-term tracker 
using tracking precision (\textbf{Pr}), tracking recall (\textbf{Re}) and tracking F-measure. 
The F-measure criterion is defined based on equation (\ref{eq-Fscore}).
\begin{equation}
\mathbf{F}(\tau_{\theta}) = 2\mathbf{Pr}(\tau_{\theta})\mathbf{Re}(\tau_{\theta}) / 
(\mathbf{Pr}(\tau_{\theta}) + \mathbf{Re}(\tau_{\theta})), 
\label{eq-Fscore}
\end{equation}
where $\tau_{\theta}$ is a given threshold. $\mathbf{Pr}({\tau _\theta })$, 
$\mathbf{Re}({\tau _\theta })$ and $\mathbf{F}({\tau _\theta })$ denote the thresholding 
precision, recall and F-measure, respectively. 
Thus, long-term tracking performance can be visualized by the tracking precision, recall, 
F-measure plots by computing these scores for all thresholds $\tau_{\theta}$.
In~\cite{VOTLT}, the \textbf{F-score} is defined as the highest score on the F-measure 
plot (i.e., taken at the tracker-specific optimal threshold), which acts as the primary role 
for ranking different trackers. 
Second, the re-detection evaluation aims to test the tracker's re-detection capability based 
on two criteria including the average number of frames required for re-detection 
(\textbf{Frames}) and the percentage of sequences with successful re-detection 
(\textbf{Success}). The detailed evaluation protocol can be found 
in the VOT2018 official report and toolbox~\cite{VOTLT}.

\begin{table}
\caption{Comparison of our tracker and $15$ competing algorithms on the VOT2018LT
dataset~\cite{VOTLT}. The best three results are marked in \textcolor{red}{\textbf{red}}, 
\textcolor{blue}{\textbf{blue}} and \textcolor{green}{\textbf{green}} bold fonts respectively.  
The trackers are ranked from top to bottom using the \textbf{F-score} measure.}
\vspace{-5mm}
\footnotesize
\begin{center}
\begin{tabular}{p{1.2cm}<{\centering}p{1.2cm}<{\centering}p{0.7cm}<{\centering}
p{0.7cm}<{\centering}p{2.6cm}<{\centering}}
\hline 
\textbf{Tracker} & \textbf{F-score} & \textbf{Pr} & \textbf{Re} & \textbf{Frames (Success)}  \\
\hline
SPLT(Ours)  &  \textbf{\textcolor[rgb]{1,0,0}{0.616}} & \textbf{\textcolor[rgb]{0,1,0}{0.633}} 
& \textbf{\textcolor[rgb]{1,0,0}{0.600}}  & 1 (100\%) \\
MBMD  &  \textbf{\textcolor[rgb]{0,0,1}{0.610}} & \textbf{\textcolor[rgb]{0,0,1}{0.634}} 
& \textbf{\textcolor[rgb]{0,0,1}{0.588}}  & 1 (100\%) \\
DaSiam\_LT  &  \textbf{\textcolor[rgb]{0,1,0}{0.607}} & 0.627 
& \textbf{\textcolor[rgb]{0,0,1}{0.588}}  & - (0\%)\\
MMLT &0.546 & 0.574 &\textbf{\textcolor[rgb]{0,1,0}{0.521}}  & 0 (100\%) \\
LTSINT & 0.536  &  0.566 & 0.510  & 2 (100\%) \\
SYT & 0.509 & 0.520 & 0.499 &  0 (43\%) \\
PTAVplus & 0.481 & 0.595 & 0.404  & 0 (11\%) \\
FuCoLoT &  0.480  & 0.539 & 0.432 & 78 (97\%) \\
SiamVGG & 0.459 & 0.552 & 0.393 & - (0\%) \\
SLT & 0.456 & 0.502 & 0.417 &  0 (100\%) \\
SiamFC & 0.433 & \textbf{\textcolor[rgb]{1,0,0}{0.636}} & 0.328 &  - (0\%) \\
SiamFCDet & 0.401 & 0.488 & 0.341 &  0 (83\%) \\
HMMTxD & 0.335 & 0.330 & 0.339 &  3 (91\%) \\
SAPKLTF & 0.323 & 0.348 & 0.300 &  - (0\%) \\
ASMS & 0.306 & 0.373 & 0.259 &  - (0\%) \\
FoT & 0.119 & 0.298 & 0.074 &  - (6\%) \\
\hline
\end{tabular}
\end{center}
\vspace{-8mm}
\label{tab-vot2018}
\end{table}

\begin{figure}[!h]
\begin{center}
\includegraphics[width=0.8\linewidth,height=0.7\linewidth]{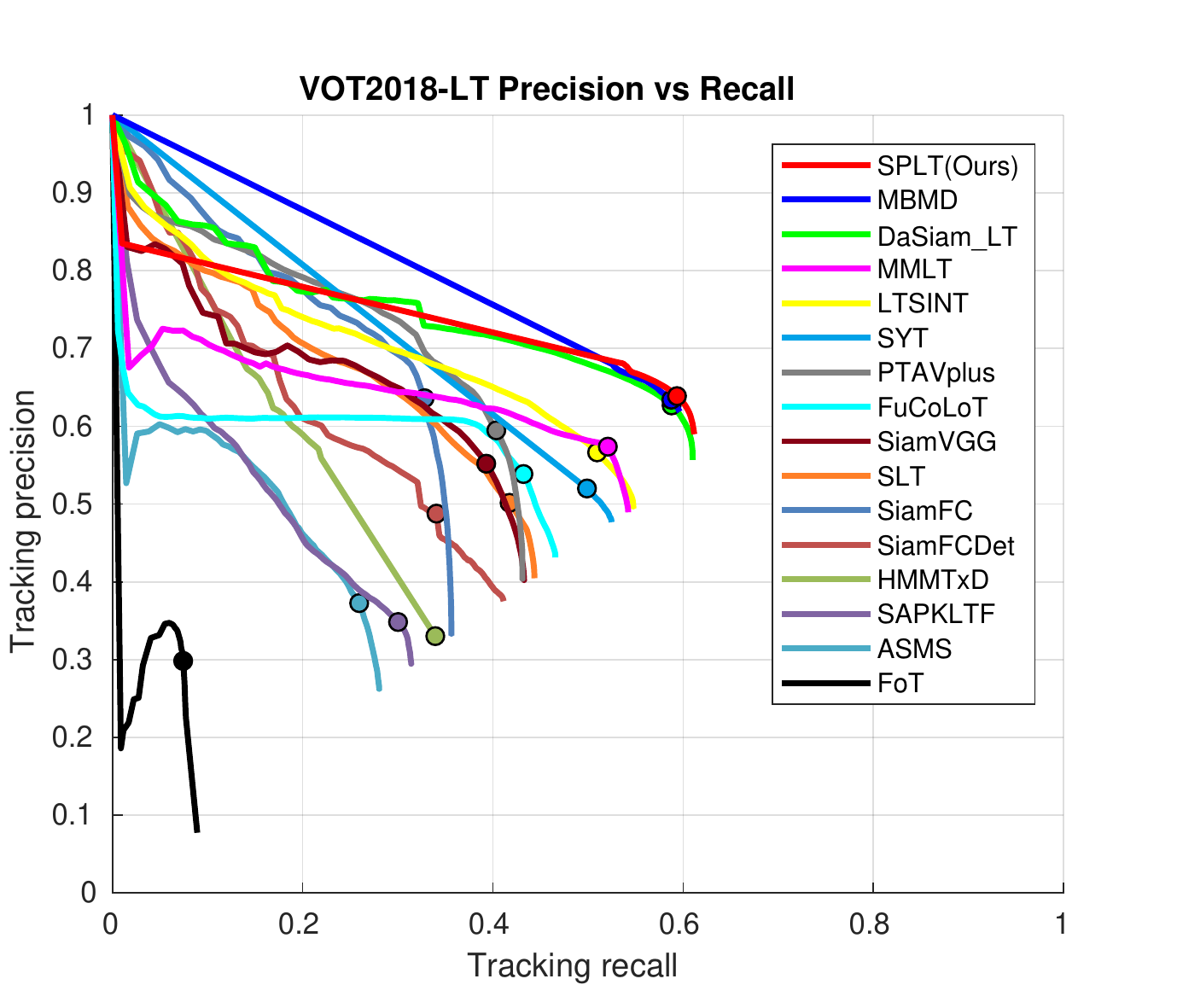}
\end{center}
\vspace{-3mm}
\caption{Average precision-recall curves of our tracker and other state-of-the-art methods 
on the VOT2018LT benchmark~\cite{VOTLT}. Different trackers are ranked based on the 
maximum of the F-Score.}
\vspace{-3mm}
\label{fig-VOT2018}
\end{figure}

Table~\ref{tab-vot2018} summarizes the comparison results of different tracking algorithms, 
from which we can see that our tracker achieves the best performance in terms of \textbf{F-score} 
and \textbf{Re} criteria while maintaining the highest re-detection success rate. 
The average precision-recall curves of our tracker and other competing ones are presented 
in Figure~\ref{fig-VOT2018}. 
Besides the proposed tracker, the MBMD and DaSiam\_LT 
methods also achieve top-ranked performance. 

\vspace{-2mm}
{\flushleft \textbf{Ours} \emph{vs} \textbf{MBMD}:} Compared with the MBMD method 
(the VOT2018 long-term challenge winner), our tracker achieves better performance in terms 
of accuracy and keeps the same re-detection capability. The most important advantage is that 
our tracker runs much faster than MBMD (25.7fps for ours and 4.4fps for MBMD in our 
experiment platform). This is mainly attributed to the proposed skimming module and the 
offline-trained verification module. Thus, we believe that our tracker could act as a new 
baseline algorithm in the VOT2018LT benchmark.
\vspace{-2mm}
{\flushleft \textbf{Ours} \emph{vs} \textbf{DaSiam\_LT}:} Compared with DaSiam\_LT 
(the second best method in the VOT2018 long-term task), the proposed tracking algorithm 
has three major strengths. First, our tracker performs better than DaSiam\_LT in terms of all 
three accuracy criteria. Second, our tracker has successfully passed the re-detection experiment 
(i.e., achieves a success rate of $100\%$), while the DaSiam\_LT method completely fails 
since its success rate is almost zero. Third, our tracker merely uses the ImageNet 
DET~\cite{ImageNet} and VID\cite{ImageNet} datasets for training, while the 
DaSiam\_LT method utilizes more than ten times training data for developing their algorithm 
(including the ImageNet DET~\cite{ImageNet}, ImageNet VID~\cite{ImageNet}, 
YTBB~\cite{Youtube} and COCO detection~\cite{COCO} datasets). 
\vspace{-2mm}
{\flushleft \textbf{Quantitative analysis on different attributes.}}
We conduct this analysis on VOT2018LT dataset and report the results in Table~\ref{tab:all_in_one}. 
Our tracker achieves top three performance in most cases. MBMD also performs 
well, but it runs much slower than ours. 

\begin{table}[h]
   \caption{\scriptsize{Quantitative analysis with respect to different attributes.
      Visual attributes: (O) Full occlusion, (V) Out-of-view, (P) Partial
      occlusion, (C) Camera motion, (F) Fast motion, (S) Scale change, (A) Aspect ratio
      change, (W) Viewpoint change, (I) Similar objects, (D) Deformable object.}}
       \vspace{-3mm}
   \begin{center}
      \resizebox{83mm}{20mm}{
         \begin{tabular}{ccccccccccccc}
            \hline
            & \bf O & \bf V & \bf P & \bf C & \bf F & \bf S & \bf A & \bf W & \bf I & \bf D \\
            \hline
            \bf Ours & \textcolor{green}{0.5666} & \textcolor{red}{0.521} & \textcolor{red}{0.5734} & \textcolor{blue}{0.6439} & 0.4520 & \textcolor{red}{0.5894} & \textcolor{red}{0.5708} & \textcolor{blue}{0.6123} & 0.5495 & \textcolor{red}{0.4991}  \\
            \bf MBMD & \textcolor{blue}{0.5693} & \textcolor{green}{0.4985} & \textcolor{blue}{0.5730} & \textcolor{green}{0.6437} & \textcolor{blue}{0.4637} & \textcolor{blue}{0.5719} & \textcolor{green}{0.5432} & \textcolor{green}{0.5780} & \textcolor{green}{0.5663} & \textcolor{blue}{0.4752}\\
            \bf DaSiam\_LT & 0.5525 & \textcolor{blue}{0.5052} & \textcolor{green}{0.5379} & \textcolor{red}{0.6515} & \textcolor{red}{0.4785} & \textcolor{green}{0.5713} & \textcolor{blue}{0.5453} & \textcolor{red}{0.6202} & \textcolor{blue}{0.5701} & 0.4570\\
            \bf MMLT & 0.5158 & 0.4823 & 0.4771 & 0.5929 & 0.4351 & 0.5188 & 0.504 & 0.5307 & 0.5213 & 0.4546 \\
            \bf LTSINT & \textcolor{red}{0.5718} & 0.4704 & 0.5358 & 0.5384 & \textcolor{green}{0.4582} & 0.5005 & 0.5123 & 0.4714 & \textcolor{red}{0.6026} & \textcolor{green}{0.4670} \\
            \bf SYT & 0.4729 & 0.4154 & 0.4508 & 0.5236 & 0.4234 & 0.4851 & 0.4255 & 0.5182 & 0.4995 & 0.3909 \\
            \bf PTAVplus & 0.4079 & 0.3143 & 0.4800 & 0.4547 & 0.1857 & 0.3599 & 0.3498 & 0.3677 & 0.3974 & 0.3738 \\
            \bf FuCoLoT & 0.4542 & 0.3378 & 0.4637 & 0.4667 & 0.2446 & 0.3959 & 0.3950 & 0.3745 & 0.4433 & 0.3828 \\
            \bf SiamVGG & 0.3265 & 0.3242 & 0.4480 & 0.4559 & 0.2195 & 0.4173 & 0.3358 & 0.4777 & 0.4445 & 0.2712 \\
            \bf SLT & 0.4067 & 0.3855 & 0.3982 & 0.5019 & 0.2871 & 0.4082 & 0.3938 & 0.4528 & 0.4110 & 0.3740 \\
            \bf SiamFC & 0.2339 & 0.2949 & 0.3523 & 0.4117 & 0.1519 & 0.3631 & 0.2759 & 0.4663 & 0.2873 & 0.2483 \\
            \bf SiamFCDet & 0.3852 & 0.3184 & 0.3170 & 0.4231 & 0.2920 & 0.3625 & 0.3640 & 0.4349 & 0.3135 & 0.3479 \\
            \bf HMMTxD & 0.2880 & 0.2584 & 0.3098 & 0.3702 & 0.2854 & 0.3362 & 0.2854 & 0.3423 & 0.3490 & 0.2705 \\
            \bf SAPKLTF & 0.2010 & 0.2230 & 0.2734 & 0.3534 & 0.1410 & 0.3243 & 0.2399 & 0.3502 & 0.3427 & 0.1598 \\
            \bf ASMS & 0.1809 & 0.2276 & 0.2437 & 0.3373 & 0.1497 & 0.2804 & 0.2294 & 0.3339 & 0.3017 & 0.1709 \\
            \bf FoT & 0.0761 & 0.1102 & 0.1280 & 0.1092 & 0.0358 & 0.1305 & 0.0989 & 0.1503 & 0.1066 & 0.0958 \\
            \hline
         \end{tabular}
      }
   \end{center}
   \vspace{-5mm}
   \label{tab:all_in_one}
\end{table}
\vspace{-4mm}
\subsection{Ablation Study}
We conduct ablation analysis to evaluate different components of our tracker using 
the VOT2018LT dataset. 
\vspace{-2mm}
{\flushleft \textbf{Effectiveness of Different Components}: }The proposed long-term 
tracking framework includes skimming (S) and perusal modules, and the perusal module 
consists of a RPN-based regressor (R) and a robust verifier (V). To evaluate the contributions 
of different components, we implement the following variants: (1) Ours (R) denotes our 
tracker merely using the SiameseRPN model to conduct local search in every frame; (2) Ours (S+R) 
stands for our tracker combining the skimming module and SiameseRPN to conduct image-wide 
re-detection in every frame; (3) Ours (R+V) represents our final tracker without the skimming 
module; and (4) Our (S+R+V) is our final skimming-perusal tracker. 

\begin{table}[!h]
\caption{Effectiveness of different components for our tracker.}
\footnotesize
\vspace{-2mm}
\begin{center}
\begin{tabular}{p{2.6cm}<{\centering}p{1.2cm}<{\centering}p{0.7cm}<{\centering}
p{0.7cm}<{\centering}p{0.7cm}<{\centering}}
\hline 
\textbf{Tracker} & \textbf{F-score} & \textbf{Pr} & \textbf{Re} & \textbf{fps}  \\
\hline
Ours (R)        &0.553 &0.561 &0.545 &34.7\\
Ours (S+R)    &0.583 &0.605 &0.563 &30.6\\
Ours (R+V)   &0.606 &0.635 &0.579 &20.0\\
Ours (S+R+V)  &0.616 &0.633 &0.600 &25.7\\
\hline
\end{tabular}
\end{center}
\vspace{-5mm}
\label{tab-ab1}
\end{table}

Table~\ref{tab-ab1} reports the results of the above-mentioned variants, and illustrates that 
all components could improve the long-term tracking performance. First, the comparison 
between Ours (R) and Ours (R+V) demonstrates that our designed verifier improves 
the long-term tracking performance by a large margin but reduces the tracking speed significantly. 
Second, the comparison between Ours (R+V) and Ours (S+R+V) illustrates that our skimming 
module could efficiently speed up long-term tracking and slightly improve the tracking performance. 
This is because the skimming module effectively selects few possible regions from a large number 
of sliding windows when the tracker conducts image-wide re-detection, which can also filter out 
some distractors and avoid sending them to the regressor. 

\begin{figure}[!h]
\begin{center}
\includegraphics[width=0.8\linewidth,height=0.55\linewidth]{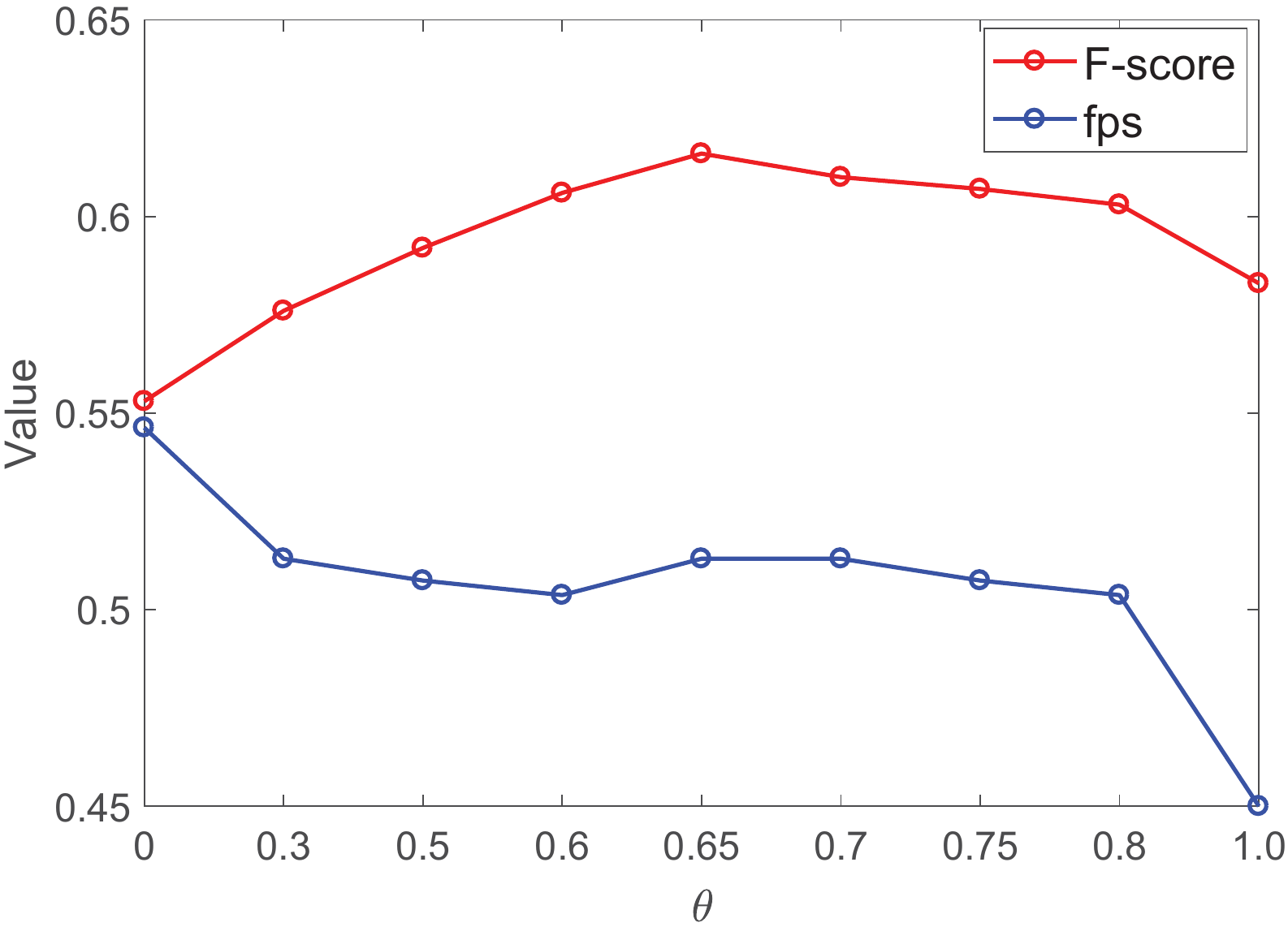}
\end{center}
\vspace{-3mm}
\caption{Effects of different $\theta$ values for dynamically switching. 
The fps values are divided by 50 for better illustrations.}
\label{fig-ab1}
\end{figure}

\vspace{-2mm}
{\flushleft \textbf{Threshold $\theta$ for Dynamically Switching}: } The threshold $\theta$ determines 
that the tracker will run on local search or global search in the next frame. A small threshold 
makes the tracker run more on the local search state. Extremely, the tracker will always track the object 
by local searching when $\theta = 0$. A large threshold treats the tracking result less reliable and makes 
the tracker exploit global search more frequently. Supposing $\theta = 1$, the tracker will always locate 
the target by image-wide re-detection. Figure~\ref{fig-ab1} illustrates the tracking accuracies and speeds 
with different $\theta$ values, which shows that our tracker achieves the best performance with a satisfactory 
speed when $\theta = 0.65$.

\begin{figure}[!h]
\begin{center}
\includegraphics[width=0.8\linewidth,height=0.55\linewidth]{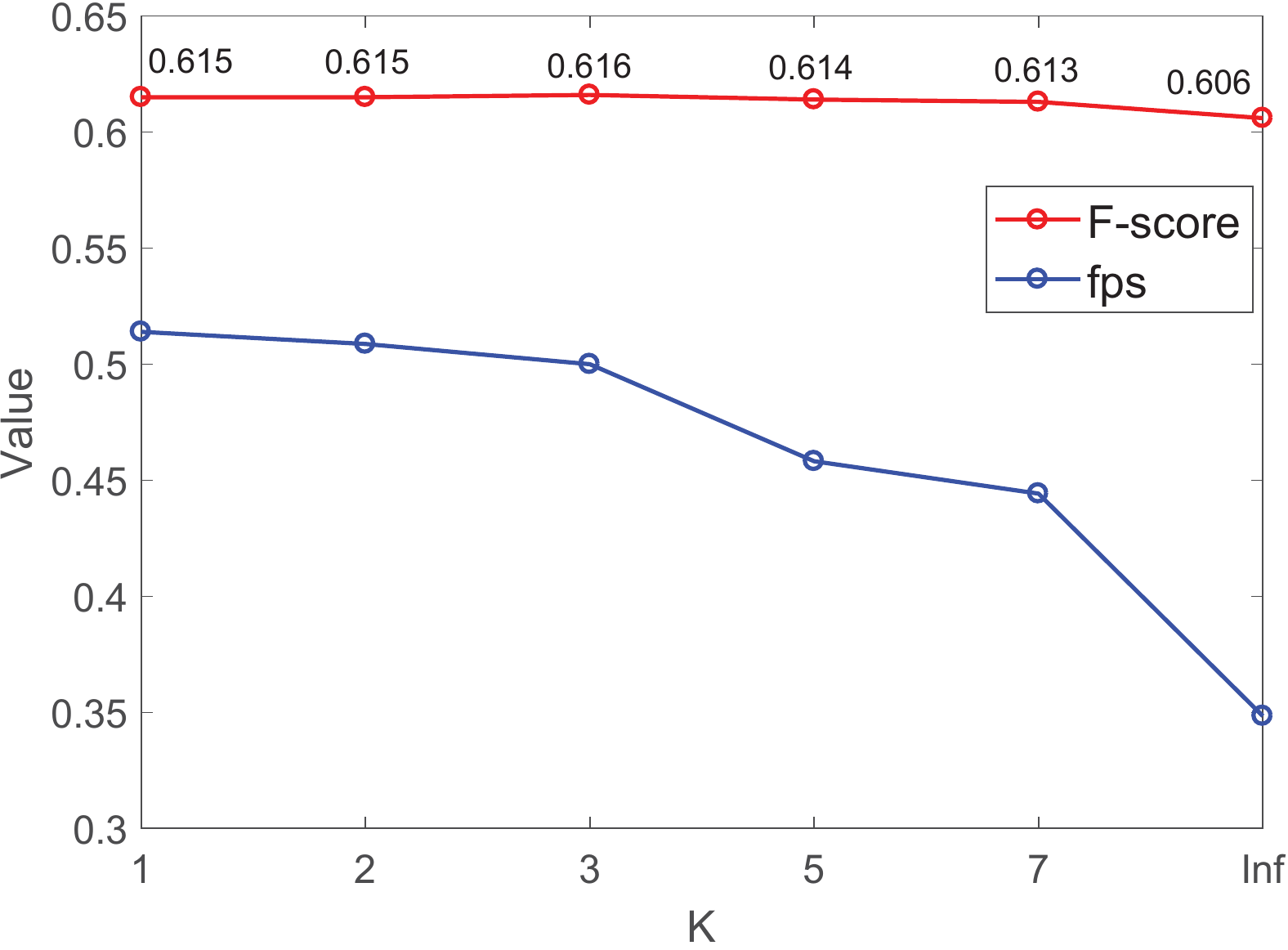}
\end{center}
\vspace{-3mm}
\caption{Effects of different $K$ values for the skimming module.
The fps values are divided by 50 for better illustrations.}
\label{fig-ab2}
\end{figure}

\vspace{-2mm}
{\flushleft \textbf{Parameter $K$ for Skimming}: } The parameter $K$ determines the 
number of possible regions being selected and sent to PRN. Figure~\ref{fig-ab2} illustrates 
that our tracker achieves the best accuracy and a satisfactory speed when $K=3$. 

\vspace{-2mm}
{\flushleft \textbf{Different Verification Networks}: }The offline-trained verification 
network aims to robust verify candidate proposals generated by SiameseRPN. We also 
investigate different kinds of networks, including MoblieNet, VGG16, ResNet50 and ResNet101. 
Table~\ref{tab-ab2} shows that our tracker with ResNet50 as verifier takes a good trade-off 
between accuracy and speed. Besides, our tracker with MoblieNet also achieves good 
performance with a faster speed. Although the ResNet101 network is more powerful than 
ResNe50 in many vision tasks, it has not achieved better results but run much slower. 
This may be attributed to the complicated network structure and limited number of training 
samples. 

\begin{table}
\caption{Comparison of different verification networks.}
\footnotesize
\begin{center}
\begin{tabular}{p{2.6cm}<{\centering}p{1.2cm}<{\centering}p{0.7cm}<{\centering}
p{0.7cm}<{\centering}p{0.7cm}<{\centering}}
\hline 
\textbf{Tracker} & \textbf{F-score} & \textbf{Pr} & \textbf{Re} & \textbf{fps}  \\
\hline
Ours(MoblieNet)    &0.597 &0.626 &0.571 &31.7\\
Ours(VGG16)         &0.596 &0.622 &0.571 &19.6\\
Ours(ResNet50)     &0.616 &0.633 &0.600 &25.7\\
Ours(ResNet101)   &0.601 &0.630 &0.571 &18.3\\
\hline
\end{tabular}
\end{center}
\vspace{-8mm}
\label{tab-ab2}
\end{table}

\subsection{Results on \textbf{OxUvA}}
The OxUvA~\cite{OxUvA} long-term dataset consists of $366$ object tracks in $337$ 
videos, which are carefully selected from the YTBB~\cite{Youtube} dataset and sparsely 
labled at a frequency of $1$Hz. 
Compared with the popular short-term tracking dataset (such as OTB2015), this dataset 
has many long-term videos (each video lasts for average 2.4 minutes) and includes severe 
out-of-view and full occlusion challenges. 
In~\cite{OxUvA}, the authors divide the OxUvA long-term dataset into two disjoint 
subsets, i.e., \emph{dev} (with $200$ tracks) and \emph{test} (with $166$ tracks) sets. 
Based on these two subsets, the OxUvA benchmark poses two challenges: 
constrained and open. For the former one, trackers can be developed merely using the 
video sequences from the OxUvA \emph{dev} set. For the open challenge, 
trackers can use any public dataset except for the YTBB \emph{validation} set since 
the OxUvA dataset is constructed upon it. 
In~\cite{OxUvA}, there exist three major criteria to evaluate the performance of different 
trackers, namely, true positive rate (\textbf{TPR}), true negative rate (\textbf{TNR}) 
and maximum geometric mean (\textbf{MaxGM}). 
\textbf{TPR} gives the fraction of \emph{present} objects that are reported \emph{present} 
and correctly located, while \textbf{TNR} measures the fraction of \emph{absent} objects that 
are reported \emph{absent}.
Then, the \textbf{MaxGM} rule (\ref{eq-MaxGM}) is defined to synthetically 
consider both \textbf{TPR} and \textbf{TNR}, which is adopted for ranking different trackers. 
For a given tracker, a larger \textbf{MaxGM} value means a better performance. 
\begin{equation}
\label{eq-MaxGM}
\begin{array}{l}
\mathbf{MaxGM} = \\
\mathop {\max }\limits_{0 \le p \le 1} \sqrt {((1 - p) \cdot \mathbf{TPR})
((1 - p) \cdot \mathbf{TNR} + p)}\\ 
\end{array}. 
\end{equation}

Until now, all trackers reported in OxUvA~\cite{OxUvA} and the recent MBMD method 
are tested using the open challenge. Thus, we also compare the proposed method with these 
trackers on the OxUvA open challenge for fair comparison. 
In this subsection, we compare the proposed tracker with the state-of-the-art MBMD method 
and ten competing algorithms reported in~\cite{OxUvA}, including LCT~\cite{LCT}, 
EBT~\cite{EBT}, TLD~\cite{TLD}, ECO-HC~\cite{ECO}, BACF~\cite{BACF}, 
Staple~\cite{Staple}, MDNet~\cite{MDNet}, SINT~\cite{SINT}, SiamFC~\cite{SiameseFC} 
and SiamFC+R~\cite{OxUvA}. 
The comparison results are presented in Table~\ref{tab-OxUvA}.
We can see that the proposed method achieves the top-ranked performance in terms 
of \textbf{MaxGM}, \textbf{TPR} and \textbf{TNR}. 
Especially, our tracker performs the best in comparison with other competing 
algorithms in terms of \textbf{MaxGM}, which is the most important metric on OxUvA. 
Compared with the VOT2018LT winner (MBMD) and the original best tracker (SiamFC+R), our 
method achieves a substantial improvement, with relative gains of $37.0\%$ and $14.3\%$ over 
\textbf{MaxGM}. 

\begin{table}[!h]
\caption{Comparisons of different tracking algorithms on the OxUvA~\cite{OxUvA} 
long-term dataset. The best three results are marked in \textcolor{red}{\textbf{red}}, 
\textcolor{blue}{\textbf{blue}} and \textcolor{green}{\textbf{green}} bold fonts 
respectively.  The trackers are ranked from top to bottom using the \textbf{MaxGM} 
measure.}
\footnotesize
\vspace{-1mm}
\begin{center}
\scalebox{1.1}{
\begin{tabular}{p{2cm}<{\centering}p{1.2cm}<{\centering}p{1.2cm}
<{\centering}p{1.2cm}<{\centering}}
\hline
\textbf{Tracker} & \textbf{MaxGM} &\textbf{TPR} & \textbf{TNR}  \\
\hline
SPLT(Ours)  &\textbf{\textcolor[rgb]{1,0,0}{0.622}} &0.498 
&\textbf{\textcolor[rgb]{0,0,1}{0.776}}  \\
MBMD  &\textbf{\textcolor[rgb]{0,0,1}{0.544}} &\textbf{\textcolor[rgb]{0,0,1}{0.609}} &0.485  \\
SiamFC+R &\textbf{\textcolor[rgb]{0,1,0}{0.454}} &0.427 &0.481 \\
TLD &0.431 & 0.208 &\textbf{\textcolor[rgb]{1,0,0}{0.895}} \\
DaSiam\_LT & 0.415 & \textbf{\textcolor[rgb]{1,0,0}{0.689}} &0 \\
LCT & 0.396 &0.292 &\textbf{\textcolor[rgb]{0,1,0}{0.537}} \\
SYT & 0.381 & \textbf{\textcolor[rgb]{0,1,0}{0.581}} &0 \\
LTSINT & 0.363 & 0.526 &0 \\
MDNet & 0.343&0.472 &0  \\
SINT & 0.326&0.426 &0  \\
ECO-HC & 0.314&0.395 & 0  \\
SiamFC & 0.313 &0.391 &0 \\
EBT & 0.283&0.321 &0\\
BACF & 0.281& 0.316&0 \\
Staple & 0.261& 0.273&0 \\
\hline
\end{tabular}}
\end{center}
\vspace{-8mm}
\label{tab-OxUvA}
\end{table}


\vspace{-3mm}
{\flushleft \textbf{Ours} \emph{vs} \textbf{Short-term Trackers}: }We first compare our tracker 
with some popular short-term trackers. ECO-HC~\cite{ECO}, BACF~\cite{BACF} and 
Staple~\cite{Staple} are three correlation-filter-based short-term trackers with high accuracies 
and fast speeds. MDNet~\cite{MDNet}, SINT~\cite{SINT} and SiamFC~\cite{SiameseFC} 
are three popular deep-learning-based short-term trackers. Compared with these methods, our 
tracker achieves a very significant improvement in terms of all three quantitative criteria. 
Table~\ref{tab-OxUvA} shows that the \textbf{TNR} values of the aforementioned short-term 
trackers are all zeros, which means that these trackers are not able to identify the 
\emph{absent} of the tracked object when it moves out of view or is fully occluded. In principle, 
these short-term methods cannot meet the requirement of the long-term tracking task. In contrast, 
our tracker includes an efficient image-wide re-detection scheme and exploits an effective 
deep-learning appearance model. 
\vspace{-2mm}
{\flushleft \textbf{Ours} \emph{vs} \textbf{Traditional Long-term Trackers}: }LCT~\cite{LCT}, 
EBT~\cite{EBT} and TLD~\cite{TLD} are three traditional long-term trackers with different 
hand-crafted features and re-detection schemes. Table~\ref{tab-OxUvA} indicates that all deep 
long-term trackers perform better than traditional ones, which means the learned deep features 
and models are also effective in the long-term tracking task. Especially, our method outperforms 
the best traditional method (TLD) by a very large margin (0.622 \emph{vs} 0.431 over 
\textbf{MaxGM}).
\vspace{-2mm}
{\flushleft \textbf{Ours} \emph{vs} \textbf{Deep Long-term Trackers}:}
SiamFC+R~\cite{OxUvA} equips the original SiamFC tracker with a 
simple re-detection scheme similar to~\cite{ODMT}.
Compared with it, our tracker exploits a more robust perusal model to 
precisely locate the tracked object, and therefore performs better than 
SiamFC+R. 
Both MBMD and our methods utilize the same SiameseRPN-based regressor but 
different verifiers. In addition, the proposed skimming module could efficiently 
not only speed up the image-wide re-detection but also filter out some distractors. 
Thus, our tracker achieves more accuracy and runs much faster than the MBMD method.
Our tracker also outperforms three VOT2018LT trackers (DaSiam\_LT, LTSINT, SYT) 
by a very large margin.
\vspace{-2mm}
\section{Conclusion}
This work presents a novel `Skimming-Perusal' tracking framework for long-term visual 
tracking. The perusal module aims to precisely locate the tracked object in a local search 
region using the offline-trained regression and verification networks. Based on the confidence 
score output by the perusal module, the tracker determines the tracked object being present 
or absent and invokes image-wide re-detection when absent. 
The skimming module focuses on efficiently selecting the most possible regions from densely 
sampled sliding windows, thereby speeding up the global search process. 
Numerous experimental results on two recent benchmarks show that our tracker achieves 
the best performance and runs at a real-time speed. 
It is worth noticing that our `Skimming-Perusal' model is a simple yet effective real-time 
long-term tracking framework. We believe that it can be acted as a new baseline for further 
researches.
\vspace{-4mm}
{\flushleft \footnotesize \textbf{Acknowledgements.}}  This paper is supported in part by National Natural 
Science Foundation of China (Nos. 61872056, 61725202, 61829102, 61751212), in part by 
the Fundamental Research Funds for the Central Universities (Nos. DUT19GJ201, 
DUT18JC30), and in part by Natural Science Foundation of Liaoning Province 
(No. 20180550673).

{\footnotesize
\bibliographystyle{ieee_fullname}
\bibliography{egbib}
}

\end{document}